\documentclass[preprint]{elsarticle}
\usepackage{lineno,hyperref}

\journal{Information Systems}

\usepackage{booktabs}
\usepackage{graphicx}
\usepackage{amsmath}
\usepackage{amssymb}
\usepackage{pifont}

\DeclareMathOperator*{\argmax}{arg\,max}%
\DeclareMathOperator*{\argmin}{arg\,min}%
\newdefinition{definition}{Definition}

\let\vec\mathbf

\makeatletter
\let\@afterindenttrue\@afterindentfalse
\makeatother

\begin{document}

\begin{frontmatter}

\title{BINet: Multi-perspective Business Process Anomaly Classification}

\author[tud]{Timo Nolle\corref{cor1}}
\ead{nolle@tk.tu-darmstadt.de}

\author[tud]{Stefan Luettgen}
\author[tud]{Alexander Seeliger}
\author[tud]{Max M\"uhlh\"auser}

\address[tud]{Telecooperation Lab, Technische Universit\"at Darmstadt, Hochschulstr. 10, 64289 Darmstadt, Germany}
\cortext[cor1]{Corresponding author}

\begin{abstract}
  In this paper, we introduce BINet, a neural network architecture for real-time multi-perspective anomaly detection in business process event logs.
  BINet is designed to handle both the control flow and the data perspective of a business process.
  Additionally, we propose a set of heuristics for setting the threshold of an anomaly detection algorithm automatically.
  We demonstrate that BINet can be used to detect anomalies in event logs not only on a case level but also on event attribute level.
  Finally, we demonstrate that a simple set of rules can be used to utilize the output of BINet for anomaly classification.
  We compare BINet to eight other state-of-the-art anomaly detection algorithms and evaluate their performance on an elaborate data corpus of 29 synthetic and 15 real-life event logs.
  BINet outperforms all other methods both on the synthetic as well as on the real-life datasets.
\end{abstract}
  
\begin{keyword}
  Business Process Management \sep Anomaly Detection \sep Artificial Process Intelligence \sep Deep Learning \sep Recurrent Neural Networks
\end{keyword}

\end{frontmatter}

\section{Introduction}\label{sec:introduction}
Anomaly detection is an important topic for today's businesses because its application areas are so manifold.
Fraud detection, intrusion detection, and outlier detection are only a few examples.
However, anomaly detection can also be applied to business process executions, for example, to clean datasets for more robust predictive analytics and robotic process automation (RPA).
Especially in RPA, anomaly detection is an integral part because the robotic agents must recognize tasks they are unable to execute not to halt the process.
Naturally, businesses are interested in anomalies within their processes because these can be indicators for inefficiencies, insufficiently trained employees, or even fraudulent activities.
Consequently, being able to detect such anomalies is of great value, for they can have an enormous impact on the economic well-being of the business.

In today's digital world, companies rely more and more on process-aware information systems (PAISs) to accelerate their processes.
Within these systems, anomaly detection should be an automatic task, and thus an utterly autonomous anomaly detection system is desirable.
Most anomaly detection algorithms still rely on a threshold that distinguishes normal from anomalous behavior.
Typically, this threshold has to be set by a user and then remains fixed during the execution.
To set the threshold adequately, the user requires a deep understanding of the underlying process, which, in the ever more complex processes of the future, is a challenging endeavor.

This paper is an extension of our previous work on BINet from 2018~\cite{nolle2018binet}.
Compared to the original publication, we have slightly simplified the architecture of BINet and present three different versions of BINet, each with different dependency modeling capabilities.
Additionally, we elaborate on the threshold heuristics proposed in~\cite{nolle2018binet} and introduce a novel set of heuristics not mentioned in the original paper.
We improved the dataset generation algorithm, which now uses an extended likelihood graph to generate causally dependent activities and event attributes.
Finally, we propose a simple rule-based classifier to distinguish different anomaly classes, solely based on the outputs of BINet.

BINet (Business Intelligence Network), is a neural network architecture that allows detecting anomalies on event attribute level.
Often, the cause of an anomaly is only represented by the value of a single attribute.
For example, a user has executed an activity without permission.
This anomaly is only represented by the user attribute of precisely this event.
Anomaly detection algorithms must work on the lowest (attribute) level, to provide the most significant benefit.
BINet has been designed to process both the control flow perspective (sequence of activities) and the data perspective (see.~\cite{van2016process}).

Due to the nature of the architecture of BINet, it can be used for ex-post analysis, but can also be deployed in a real-time setting to detect anomalies at runtime.
Being able to detect anomalies at runtime is crucial because otherwise no countermeasures can be initiated in time.
BINet can be trained during the execution of the process and therefore can adapt to concept drift.
If unseen attribute values occur during the training, the network can be altered and retrained on the historical data to include the new attribute value in the future.
Dealing with concept drift is also important since most business processes are flexible systems.
BINet can detect point anomalies as well as contextual anomalies (see~\cite{han2011data}).

BINet works under the following assumptions.
\begin{enumerate}
  \item No domain knowledge about the process
  \item No clean dataset (i.e., the dataset contains anomalous examples)
  \item No reference model
  \item No labels (i.e., no knowledge about anomalies)
  \item No manual threshold
\end{enumerate}

In the context of business processes, an anomaly is defined as a deviation from a defined behavior, i.e., the business process. 
An anomaly is an event that does not typically occur as a consequence of preceding events; specifically, their order and the combination of attributes.
Anomalies that are attributed to the order of activities (e.g., two activities are executed in the wrong order) are called control flow anomalies.
Anomalies that are attributed to the event attributes (e.g., a user has executed an activity without permission) are called data anomalies.

We compare BINet to eight state-of-the-art anomaly detection methods and evaluate on a comprehensive dataset of 29 synthetic logs and 15 real-life logs, using artificial anomalies. This work contains five main contributions.
\begin{enumerate}
  \item BINet neural network architecture
  \item Automatic threshold heuristics
  \item Generation algorithm for synthetic event logs
  \item Comprehensive evaluation of state-of-the-art methods
  \item Method to classify anomalies based on outputs of BINet
\end{enumerate}

\begin{figure}[t]
  \centering
  \includegraphics[width=1.0\linewidth]{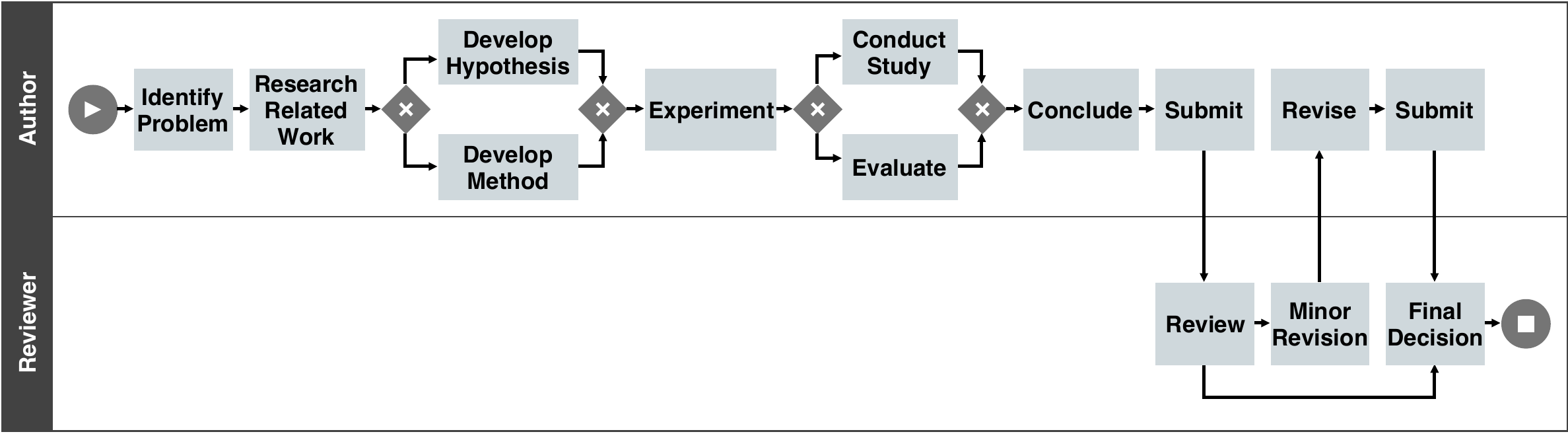}
  \caption{A simple paper submission process which is used as an example throughout the paper}
  \label{fig:process}
\end{figure}

Throughout this paper, we use a simple paper submission process as the primary example to illustrate concepts, methods, and results.
The process model in Fig.~\ref{fig:process} describes the creation of a scientific paper.
Note that the process includes the peer review process, which is executed by a reviewer, whereas the paper is conceptualized and compiled by an author.
We return to this process in Sec.~\ref{sec:datasets} when describing the dataset generation.

\section{Related Work}\label{sec:relatedwork}
In the field of process~mining~\cite{van2016process}, it is popular to use discovery algorithms to mine a process model from an event log and then use conformance checking to detect anomalous behavior~\cite{wen2007mining,Bezerra2009Anomaly,bezerra2008anomaly}.
However, the proposed methods do not utilize the event attributes, and therefore cannot be used to detect anomalies on attribute level.

A more recent publication proposes the use of likelihood graphs to analyze business process behavior~\cite{bohmer2016multi}.
This method includes important characteristics of the process itself by including the event attributes as part of an extended likelihood graph.
However, this method relies on a discrete order in which the attributes are connected to the graph, which may introduce a bias towards certain attributes.
Furthermore, the same activities are mapped to the same node in the likelihood graph, thereby assigning a single probability distribution to each activity.
In other words, control flow dependencies cannot be modeled by the likelihood graph, because the probability distribution of attributes following an activity does not depend on the history of events.

The main drawback of this method is that it uses the initial log to build the likelihood graph, and therefore no case of the original log is classified as anomalous.
This is related to the method the authors chose to determine a threshold for the anomaly detection task.
We address this caveat again in Sec.~\ref{sec:evaluation}.
Nevertheless, the notion of the likelihood graph inspired the generation method for synthetic event logs in Sec.~\ref{sec:datasets}.

A review of traditional anomaly detection methodology can be found in~\cite{pimentel2014review}.
Here, the authors describe and compare many methods that have been proposed over the last decades.
Another elaborate summary of anomaly detection in discrete sequences is given by Chandola  et\,al.\ in~\cite{chandola2012survey}.
The authors differentiate between five different basic methods for novelty detection: probabilistic, distance-based, reconstruction-based, domain-based, and information-theoretic novelty detection.

Probabilistic approaches estimate the probability distribution of the normal class and thus can detect anomalies as they come from a different distribution.
An important probabilistic technique is the sliding window approach~\cite{warrender1999detecting}.
In window-based anomaly detection, an anomaly score is assigned to each window in a sequence.
Then the anomaly score of the sequence can be inferred by aggregating the window anomaly scores.
Recently, Wressnegger et\,al.\ used this approach for intrusion detection and gave an elaborate evaluation in~\cite{wressnegger2013acloselook}.
While being inexpensive and easy to implement, sliding window approaches show a robust performance in finding anomalies in sequential data, especially within short regions~\cite{chandola2012survey}.

Distance-based novelty detection does not require a clean dataset, yet it is only partly applicable to process cases, as anomalous cases are usually very similar to normal ones.
A popular distance-based approach is the one-class support vector machine (OC-SVM).
Sch\"olkopf et\,al.~\cite{scholkopf1999support} first used support vector machines~\cite{cortes1995support} for anomaly detection.

Reconstruction-based novelty detection (e.g., neural networks) is based on the idea to train a model that can reconstruct normal behavior but fails to do so with anomalous behavior. Therefore, the reconstruction error can be used to detect anomalies~\cite{Japkowicz2001}.
This approach has successfully been used for the detection of control flow anomalies~\cite{nolle2016unsupervised} as well as data flow anomalies~\cite{nolle2018analyzing} in event logs of PAISs.

Domain-based novelty detection requires domain knowledge, which violates our assumption of no domain knowledge about the process.
Information-theoretic novelty detection defines anomalies as the examples that influence an information measure (e.g., entropy) on the whole dataset the most.
Iteratively removing the data with the highest impact yields a cleaned dataset and thus a set of anomalies.

The core of BINet is a recurrent neural network, trained to predict the next event and its attributes.
The architecture is influenced by the works of Evermann~\cite{evermann2016deep,evermann2017predicting} and Tax~\cite{tax2017predictive}, who utilized long short-term memory~\cite{hochreiter1997long} (LSTM) networks for next event prediction, demonstrating their utility.
LSTMs have been used for anomaly detection in different contexts like acoustic novelty detection~\cite{marchi2015novel} and predictive maintenance~\cite{malhotra2016lstm}.
These applications mainly focus on the detection of anomalies in time series and not, like BINet, on multi-perspective anomaly detection in discrete sequences of events.

The novelty of BINet lies in the tailored architecture for business processes, including the control flow and data perspective, the scoring function to assign anomaly scores, and the automatic threshold heuristic.
It is a universally applicable method for anomaly detection both in the control flow and the data perspective of business process event logs.
Furthermore, BINet fulfills all the requirements connected to the assumptions above.
Lastly, BINet can handle multiple event attributes and model causal dependencies between control flow and data perspective, as well dependencies between event attributes.
This combination is, to the best of our knowledge, novel to the field.

\section{Datasets}\label{sec:datasets}
As a basis for the following sections, we first want to define the terms case, event, log, and attribute.
A log consists of cases, each of which consists of events executed within a process.
Each event is defined by an activity name and its attributes, e.g., a user who executed the event.
We use a nomenclature adapted from~\cite{van2016process}.
\begin{definition}{Case, Event, Log, Attribute.}
  Let $\mathcal{C}$ be the set of all cases, and $\mathcal{E}$ be the set of all events.
  The event sequence of a case $c \in \mathcal{C}$, denoted by $\hat{c}$, is defined as $\hat{c} \in \mathcal{E}^{*}$, where $\mathcal{E}^{*}$ is the set of all sequences over $\mathcal{E}$.
  An event log is a set of cases $\mathcal{L} \subseteq \mathcal{C}$.
  Let $\mathcal{A}$ be a set of attributes and $\mathcal{V}$ be a set of attribute values, where $\mathcal{V}_{a}$ is the set of possible values for the attribute $a \in \mathcal{A}$.
  Note that $|\hat{c}|$ is the number of events in case $c$, $|\mathcal{L}|$ is the number of cases in $\mathcal{L}$, and $|\mathcal{A}|$ is the number of event attributes.
\end{definition}

\subsection{Synthetic Dataset Generation}
To evaluate our method, we generated synthetic event logs from random process models of different complexities.
We used PLG2~\cite{burattin2015plg2} to generate six random process models.
The complexity of the models varies in the number of activities, breadth, and width.
We also use a handmade procurement process model called P2P as in~\cite{nolle2018binet}.
For demonstrative purposes, we also include the Paper process from Fig.~\ref{fig:process} in the datasets since it features human readable activities.

\begin{figure}[t]
  \centering
  \includegraphics[width=1.0\linewidth]{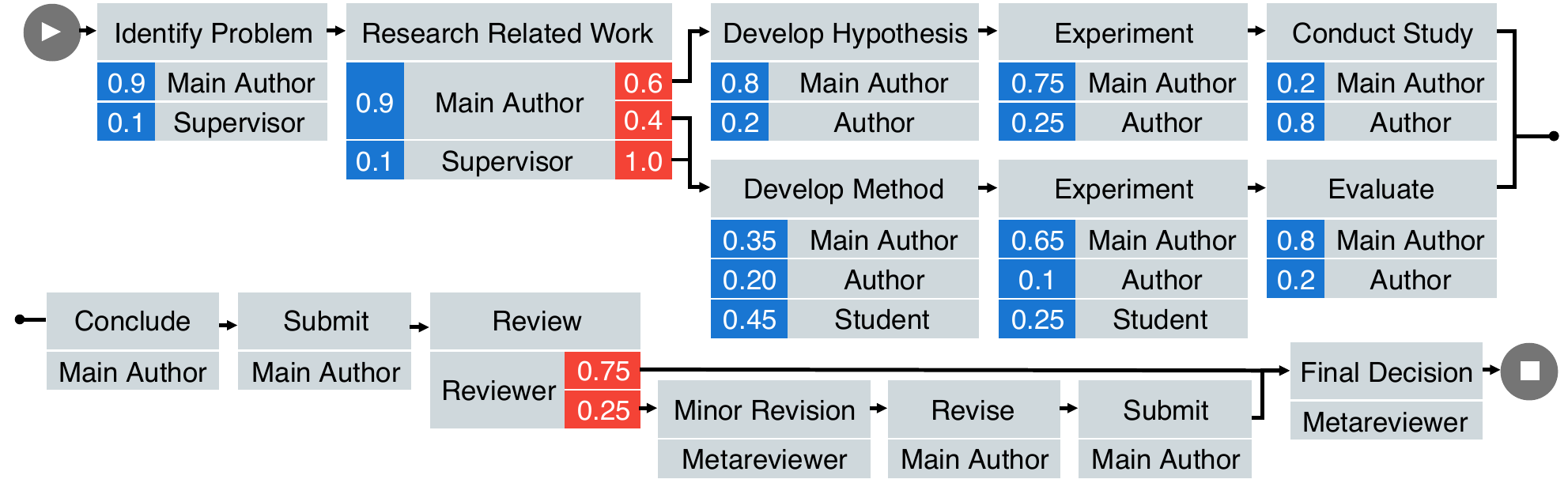}
   \caption{A likelihood graph with user attribute; 1.0 probabilities omitted for simplicity}
  \label{fig:likelihood}
\end{figure}

We adopt the notion of the extended likelihood graph (cf.~\cite{bohmer2016multi}) to generate causally dependent event attributes.
For each activity in the process from Fig.~\ref{fig:process}, we create a group of possible users allowed to execute the activity.
Additionally, we assign different probabilities to each user.
Figure~\ref{fig:likelihood} demonstrates the final result.

Note that the \textit{Experiment} activity appears twice in the likelihood graph.
This is to introduce a long-term control flow dependency. 
That is, \textit{Conduct Study} always eventually follows \textit{Develop Hypothesis}, and never eventually follows \textit{Develop Method}.
Note also that the user group, as well as the corresponding probabilities, are different.

This method can easily be extended to generate more than one event attribute.
For simplicity, the visualization in Fig.~\ref{fig:likelihood} uses a table-like structure to depict the activities and the possible users.
In reality, this is implemented as a directed graph, where each possible user is a direct follower of the activity, with the probabilities being the edge weights.
Hence, we can add more attributes by adding additional successors to each user and so on.
Hereby, causal dependencies between event attributes can be modeled (e.g., \textit{Main Author} only works Mondays and Tuesdays).

We have described long-term control flow dependencies as well as data dependencies.
Data to control flow dependencies are also possible, as in activity \textit{Research Related Work}.
\textit{Develop Method} always directly follows \textit{Research Related Work} if \textit{Student} is the user.

Now, we can generate event logs by using a random-walk through the likelihood graph, complying with the transition probabilities, generating activities and attributes along the way.
We implemented the generation algorithm so that all these dependencies can be controlled by parameters and the event attributes are automatically generated.
Please refer to the code repository\footnote{https://github.com/tnolle/binet}, and specifically the notebooks section, for a detailed description of the algorithm as well as examples.

In addition to the synthetic logs, we also use the real-life event logs from the Business Process Intelligence Challenge (BPIC): 
BPIC12\footnote{http://www.win.tue.nl/bpi/doku.php?id=2012:challenge}, 
BPIC13\footnote{http://www.win.tue.nl/bpi/doku.php?id=2013:challenge}, 
BPIC15\footnote{http://www.win.tue.nl/bpi/doku.php?id=2015:challenge}, and 
BPIC17\footnote{http://www.win.tue.nl/bpi/doku.php?id=2017:challenge}.
Furthermore, we use a set of 4 event logs (Anonymous) from real-life procurement processes provided by a consulting company.

\subsection{Artificial Anomalies}
Like Bezerra~\cite{bezerra2013algorithms} and B\"ohmer~\cite{bohmer2016multi}, we apply artificial anomalies to the event logs, altering $30$ percent of all cases.
Inspired by the anomaly types used in~\cite{bezerra2013algorithms, bohmer2016multi} (\textit{Skip}, \textit{Insert}, and \textit{Switch}) we identified more elaborate  anomalies that frequently occur in real business processes.
These anomalies are defined as follows.
\begin{enumerate}
  \item \textit{Skip}: A necessary sequence of up to 3 events has been skipped
  \item \textit{Insert}: Up to 3 random activities have been inserted
  \item \textit{Rework}: A sequence of up to 3 events has been executed a second time
  \item \textit{Early}: A sequence of up to 2 events has been executed too early, and hence are skipped later in the case
  \item \textit{Late}: A sequence of up to 2 events has been executed too late, and hence are skipped earlier in the case
  \item \textit{Attribute}: An incorrect attribute value has been set in up to 3 events
\end{enumerate}

Notice that we do apply the artificial anomalies to the real-life event logs as well, knowing that they likely already contain natural anomalies which are not labeled.
Thereby, we can measure the performance of the algorithms on the real-life logs to demonstrate feasibility while using the synthetic logs to evaluate accuracy.

\begin{figure}[t]
  \centering
  \includegraphics[width=1.0\linewidth]{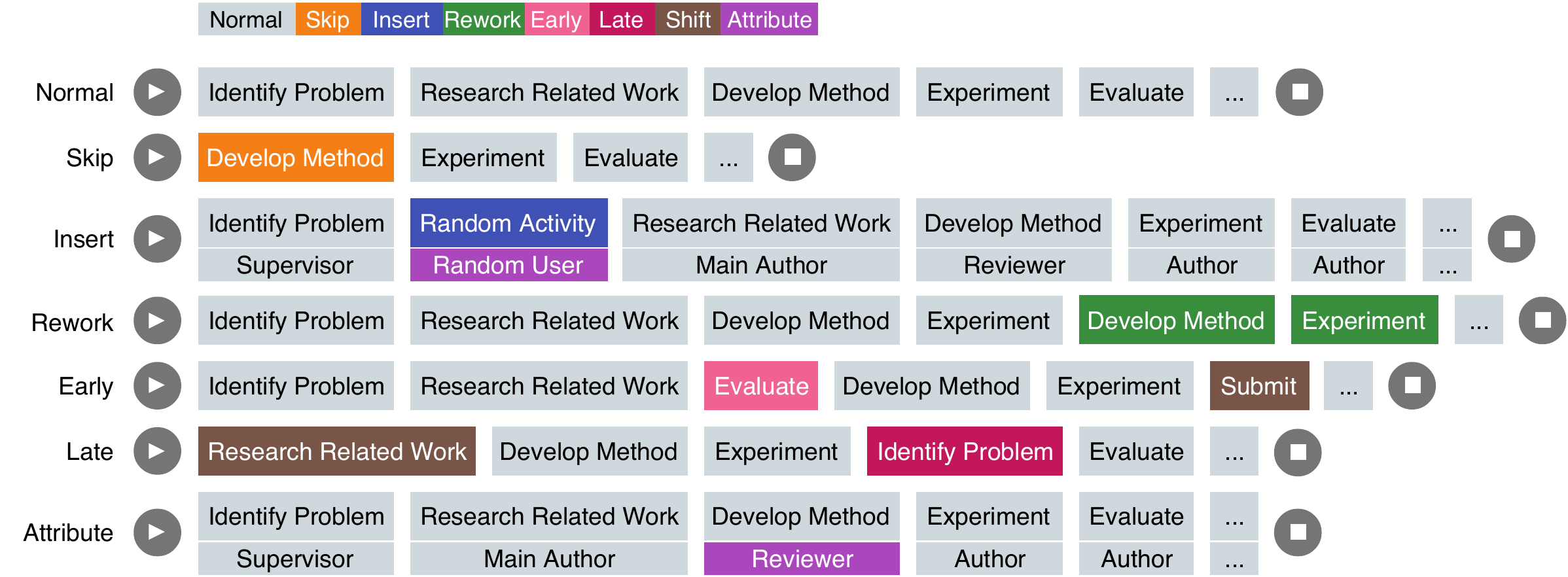}
  \caption{Anomalies applied to cases of the paper submission process} 
  \label{fig:anomalies}
\end{figure}

As indicated in Fig.~\ref{fig:anomalies} we can gather a ground truth dataset by marking the attributes with their respective anomaly types.
Note that we introduce a \textit{Shift} anomaly type, which is used to indicate the place where an \textit{Early} or \textit{Late} event used to be.
Essentially this is equivalent to a \textit{Skip}; however, we want to differentiate these two cases.
This becomes important later on.

Notice that we insert a random event in case of \textit{Insert}, i.e., the activity name does not come from the original process.
This is to prevent a random insert from resembling, say, \textit{Rework}, which is possible when randomly choosing an activity from the original process.
Attribute values of inserted events are set in the same fashion.
When applying an \textit{Attribute} anomaly, we randomly select an attribute value from the likelihood graph that is not a direct successor of the last node.

So, we created datasets with ground truth data on attribute level.
For the anomaly detection task, these labels are mapped to either \textit{Normal} or \textit{Anomaly}, thus creating a binary classification problem.
The ground truth data can easily be adapted to case level by the following rule: A case is anomalous if any of the attributes in its events are anomalous.

We generated 4 likelihood graphs for each synthetic process model with different numbers of attributes, different transition probabilities, and dependencies.
Then, we sampled logs from these likelihood graphs, resulting in 28 synthetic logs (29, including the Paper dataset).
Together with BPIC12, BPIC13, BPIC15, BPIC17, and Anonymous, the corpus consists of 44 event logs.
We refer to the datasets by their names as defined in Table~\ref{tab:datasets}, which gives a detailed overview of the corpus.
\begin{table}[t]
  \centering
  \label{tab:datasets}
  \resizebox{\columnwidth}{!}{%
  \begin{tabular}{lcccccc}
      \toprule
      Name.     & \#Logs & \#Activities & \#Cases      & \#Events         & \#Attr.  & \#Attr. Values \\ \midrule
      Paper     & 1      & 27           & 5K           &  66K             & 1        & 13             \\
      P2P       & 4      & 27           & 5K           & 48K--53K        & 1--4     & 13--386        \\
      Small     & 4      & 41           & 5K           & 53K--57K        & 1--4     & 13--360        \\
      Medium    & 4      & 65           & 5K           & 39K--42K        & 1--4     & 13--398        \\
      Large     & 4      & 85           & 5K           & 61K--68K        & 1--4     & 13--398        \\
      Huge      & 4      & 109          & 5K           & 47K--53K        & 1--4     & 13--420        \\
      Gigantic  & 4      & 154--157     & 5K           & 38K--42K        & 1--4     & 13--409        \\
      Wide      & 4      & 68--69       & 5K           & 39K--42K        & 1--4     & 13--382        \\ 
      \midrule
      BPIC12    & 1      & 73           & 13K          & 290K            & 0        & 0              \\
      BPIC13    & 3      & 11--27       & 0.8K--7.5K   & 4K--81K         & 2--4     & 23--1.8K       \\
      BPIC15    & 5      & 422--486     & 0.8K--1.4K   & 46K--62K        & 2--3     & 23--481        \\
      BPIC17    & 2      & 17--53       & 31K--43K     & 284K--1.2M      & 1        & 289--299       \\
      \midrule
      Anonymous & 4      & 19--37       & 968--17K     & 6.9K--82K      & 1        & 160--362 \\
      \bottomrule
  \end{tabular}
  }
  \caption{Overview showing dataset information}
\end{table}

\section{Method}\label{sec:method}
In this section, we describe the BINet architecture and how it is utilized for anomaly detection.

\subsection{Preprocessing}\label{sec:preprocessing}
Due to the mathematical nature of neural networks, we must transform the logs into a numerical representation.
To accomplish this, we encode all nominal attribute values by using an integer encoding.
An integer encoding is a mapping $\mathcal{I}_a: \mathcal{V}_a \rightarrow \mathbb{N}$ of all possible attribute values for an attribute $a$ to a unique positive integer.
The integer encoding is applied to all attributes of the log, including the activity name.

Now, event logs can be represented as third-order tensors.
Each event $e$ is a first-order tensor $\vec{e} \in \mathbb{R}^{A}$, with $A = |\mathcal{A}|$, the first attribute always being the activity name, representing the control flow perspective.
Hence, an event is defined by its activity name and the event attributes.
Each case is then represented as a second-order tensor $\vec{C} \in \mathbb{R}^{E \times A}$, with $E = \max_{c \in \mathcal{L}} |\hat{c}|$, being the maximum case length of all cases in the log $\mathcal{L}$.
To force all cases to have the same size, we pad all shorter cases with event tensors only containing zeros, which we call padding events (these are ignored by the neural network).

The log $\mathcal{L}$ can now be represented as a third-order tensor $\vec{L} \in \mathbb{R}^{C \times E \times A}$, with $C = |\mathcal{L}|$, the number of cases in log $\mathcal{L}$.
Using matrix index notation, we can now obtain the second attribute of the third event in the ninth case with $\vec{L}_{9,3,2}$.
We can also obtain all the second attributes of the third event by $\vec{L}_{:,3,2}$, using ``:'' to denote the cross-section of tensor $\vec{L}$ along the case axis.
Likewise, we can obtain all the second attributes of case nine by $\vec{L}_{9,:,2}$.
Thus, we can define a preprocessor as follows:
\begin{definition}{Preprocessor}
  Let $C$, $E$, and $A$ be defined as above, then a preprocessor is a mapping $\mathcal{P}: \mathcal{L} \rightarrow \mathbb{R}^{C \times E \times A}$.
\end{definition}

The preprocessor $\mathcal{P}$ encodes all attribute values and then transform the log $\mathcal{L}$ into its tensor representation.
In the following, we refer to the preprocessed log $\mathcal{L}$ by $\vec{F}$ (features), with $\vec{F} = \mathcal{P}(\mathcal{L})$.

\subsection{BINet Architecture}
BINet is based on a neural network architecture that is trained to predict the attributes of the next event.
To model the sequential nature of event log data, the core of BINet is a recurrent neural network, using a Gated Recurrent Unit (GRU)~\cite{cho2014learning}, an alternative to the popular long short-term memory (LSTM)~\cite{hochreiter1997long}.

BINet processes the distinct sequence of events for each case.
For each event, BINet has to predict the next event based on the history of events in the case.
Thus, BINet is a sequence-to-sequence recurrent neural network.

It is important to understand that for each prediction, BINet has a recollection of not only the last event in the sequence but all of the events.
The internal state of the GRU units changes with each new event and resembles a latent representation of the sequence of events up until that point.
For each new case, these internal states are reset.

\begin{figure}[t]
  \centering
  \includegraphics[width=0.8\linewidth]{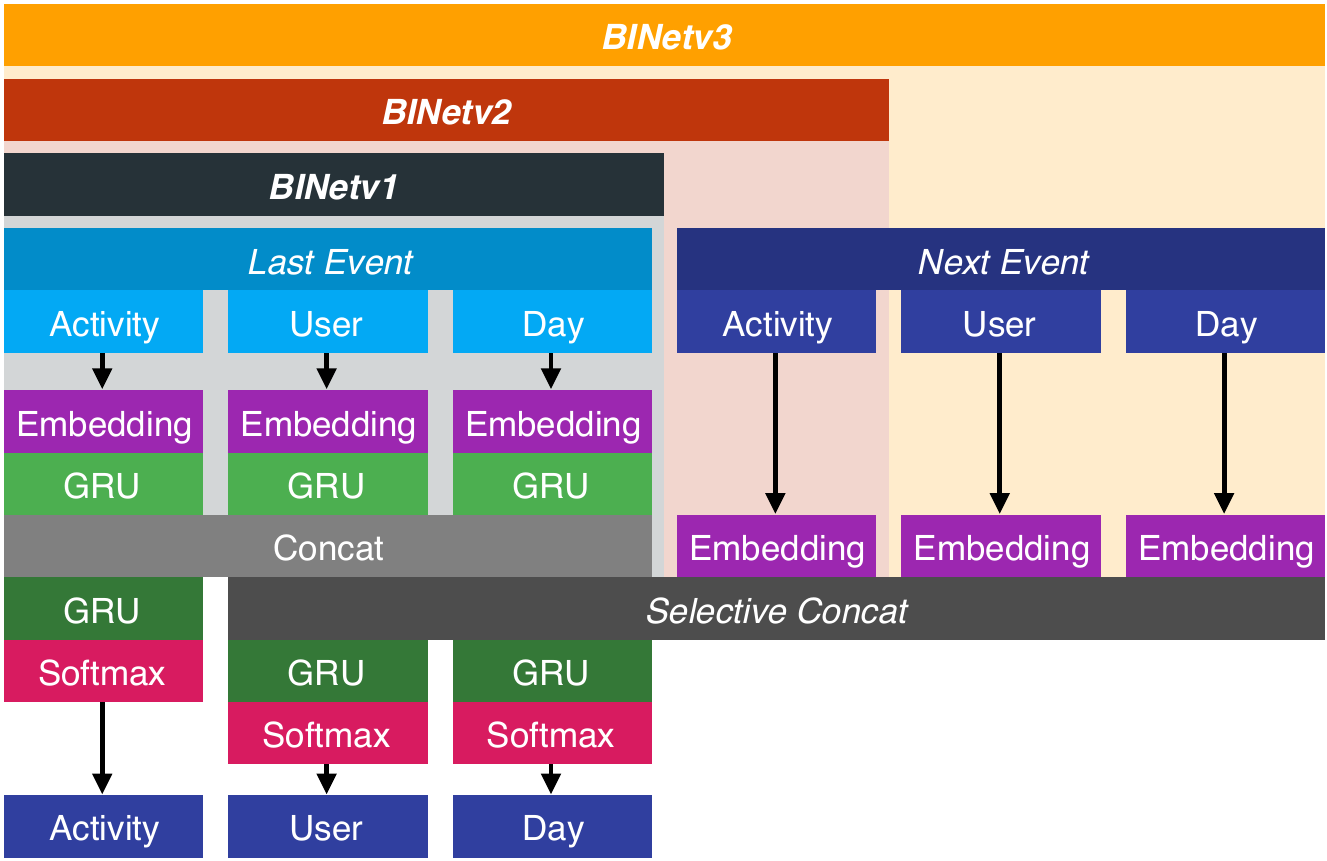}
  \caption{BINet architectures for a log with two event attributes, \textit{User} and \textit{Day}; the three versions of BINet differ only in the inputs they receive}
  \label{fig:architecture}
\end{figure}

Figure~\ref{fig:architecture} shows the internal architecture of BINet.
We propose three versions of BINet (BINetv1, BINetv2, and BINetv3).
These versions differ in their capability of modeling causal dependencies based on the inputs they receive.
BINet consists of dedicated encoder GRUs (light green) for each input attribute (light blue) of the last event.
These GRUs are responsible for creating a latent representation of the complete history of a single attribute.
Note that each attribute is fed through an embedding layer to reduce the input dimension (see~\cite{mikolov2013efficient, dekoninck2018trace2vec}).

The counterpart to the encoder GRUs are the decoder GRUs (dark green) which receive as input a concatenation of attribute representations.
These GRUs are responsible for combining the control flow and the data perspective, and hence to create a latent representation of the complete history of events that is meaningful to the respective next attribute prediction in the next layer.
For prediction, BINet uses a softmax output layer (pink) for each next attribute.
A softmax layer outputs a probability distribution over all possible values of the respective attribute.

In its simplest form, BINet predicts the next activity and all next attributes solely based on past events.
This architecture is called BINetv1 (black).
However, there likely exist causal dependencies between the activity and the corresponding attribute values for that activity.

To model this dependency, we propose a second architecture, BINetv2 (red), which in addition to the input of BINetv1, gets access to the activity of the next event.
This is to condition the attribute decoders onto the actual next activity, as opposed to inferring the next activity from the states of the encoders.
Using the BINetv2 architecture, we can now model control flow to data dependencies.

There also likely exist dependencies between attributes (i.e., certain users only work certain days of the week).
With BINetv1 or BINetv2, these dependencies are not modeled because the attributes are treated as though they were independent.
To address this, the last architecture we propose is BINetv3 (orange), which gets access to the complete next event.
Hence, BINetv3 can model data to data dependencies.

Note that the selective concatenation layer (dark grey) is special in the sense that it does not allow information to flow from one of the next event inputs to the respective next attribute decoder GRU.
We must prevent this because otherwise, BINetv3 can simply predict correctly by observing the attributes in the next event.
This problem does not arise with BINetv1 and BINetv2, because the activity GRU has no direct connection to the next event, and the user and day GRUs only have access to the next activity.
Apart from the encoder-decoder structure, BINetv2 is essentially equivalent to the original BINet architecture from~\cite{nolle2018binet}.

We want to elaborate on the differences in the BINet versions by referring back to the paper submission process from Fig.~\ref{fig:likelihood}.
Suppose the last activity input to BINet is \textit{Research Related Work} and the user was \textit{Main Author}.
The activity output should now give a probability of approximately 60 percent to \textit{Develop Hypothesis}.
In case of BINetv1, however, the user output does not match the 80 percent for \textit{Main Author} and the 20 percent for \textit{Author}, because the respective decoders also have to take into account the other 40 percent of not going to \textit{Develop Hypothesis}, and hence output a higher probability for \textit{Student}.
BINetv2 does not suffer from this problem, because it knows for certain that \textit{Develop Hypothesis} is the next activity (because of the next activity input), and thus can learn the probabilities appropriately.

To demonstrate the advantage of BINetv3, we have to imagine a third weekday attribute as part of the extended likelihood graph.
Suppose for a given activity \textit{Main Author} works only on Fridays, and \textit{Author} works from Monday until Thursday.
BINetv3 can correctly predict that if the weekday is Friday, the user must be \textit{Main Author}, whereas BINetv2 is not.
Likewise, BINetv3 can infer that if the user is \textit{Main Author}, the day must be Friday.

For BINetv1, activity, user, and day are entirely independent, for BINetv2, user and day are dependent on the activity, and for BINetv3, user is dependent on activity and day, whereas day is dependent on activity and user.
Our implementation of BINet theoretically allows for any number of events and attributes.

\subsection{Calculating Anomaly Scores}
After the initial training phase, BINet can now be used for anomaly detection.
This is based on the assumption that BINet assigns a lower probability to an anomalous attribute than a normal attribute.

The last step of the anomaly detection process is the scoring of the events.
Therefore, we use a scoring function in the last layer of the architecture.
This scoring function for an attribute $a$ receives as input the output of the softmax layer for $a$, that is, a probability distribution $\vec{p}_a$, and the actual value of the attribute, $v$.

\begin{figure}[t]
  \centering
  \includegraphics[width=0.8\linewidth]{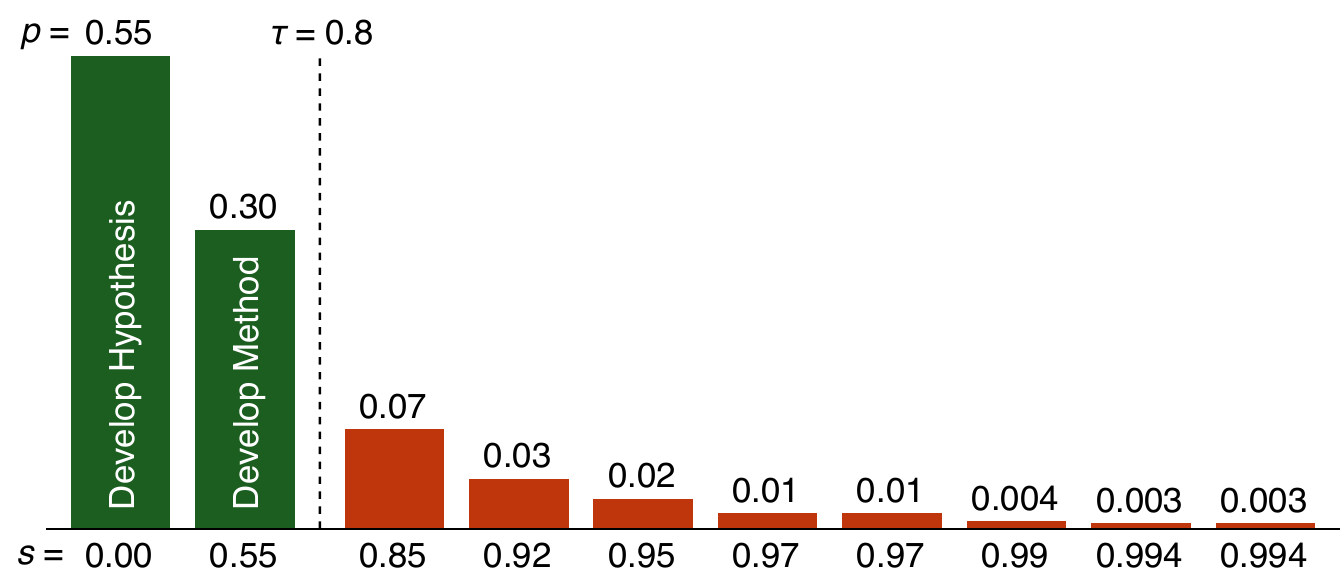}
  \caption{Output of the activity softmax layer after reading activity \textit{Research Related Work} and user \textit{Main Author}}
  \label{fig:scores}
\end{figure}

Using the example above (the last activity being \textit{Research Related Work} and the user being \textit{Main Author}), the output of the activity softmax layer might look as depicted in Fig.~\ref{fig:scores}.
The probability $p$ for \textit{Develop Hypothesis} is $0.55$, and the probability for \textit{Develop Method} is $0.30$.
Note that BINet gives a high probability for the two correct next activities with respect to the paper process.

We can now define the anomaly score for a possible attribute value $v$ as the sum of all probabilities $p$ of the probability distribution tensor $\vec{p}$ greater than the probability assigned to $v$, $p_v$.
The scoring function $\sigma$ is therefore defined as follows, with $\vec{p}_i$ being $i$-th probability.
\begin{equation*}
  \sigma(\vec{p}, p_v) = \sum_{\vec{p}_i > p_v} \vec{p}_i
\end{equation*}
Figure~\ref{fig:scores} also shows the resulting anomaly scores, $s$, for each possible activity.
Intuitively, an anomaly score of $0.55$ indicates that the probability of an attribute value lies within the top 55 percent (plus a small margin) confidence interval of BINet.
Thus, we can set a threshold as indicated in Fig.~\ref{fig:scores} ($\tau = 0.8$), to flag all values as normal that lie within the first 80 percent of BINet's confidence interval.

The scoring function $\sigma$ is applied to each softmax output of BINet, transforming the probability distribution tensor into a scalar anomaly score.
We can now obtain the anomaly scores tensor $S$ by applying BINet to the feature tensor $\vec{F}$,
\begin{equation*}
  \vec{S} = \left(s_{ijk}\right) \in \mathbb{R}^{C \times E \times A} = \text{BINet}(\vec{F}),
\end{equation*}
mapping an anomaly score to each attribute in each event in each case.
The anomaly score for attributes of padding events is always 0.

\subsection{Training}
BINet is trained without the scoring function.
The GRU units are trained in sequence to sequence fashion.
With each event that is fed in, the network is trained to predict the attributes of the next event.
We train BINet with a GRU size of $2E$ (two times the maximum case length), on mini batches of size $500$ for 20 epochs using the Adam optimizer with the parameters stated in the original paper~\cite{kingma2014adam}.
Additionally, we use batch normalization~\cite{ioffe2015batch} after each GRU to counteract overfitting.

\subsection{Detection}
An anomaly detector only outputs anomaly scores.
We need to define a function that maps anomaly scores to a label $l \in \{0, 1\}$, $0$ indicating normal and $1$ indicating anomalous, by applying a threshold $\tau$.
Whenever an anomaly score for an attribute is greater than $\tau$, this attribute is flagged as anomalous.
Therefore, we define a threshold function $\theta$, with inputs $\vec{S}$ and $\tau \in \mathbb{R}$.
\begin{equation*}
  \theta(\vec{S}, \tau) = \vec{Y}_{ijk} = 
  \begin{cases}
    1 & \text{if } \vec{S}_{ijk} > \tau \\
    0 & \text{otherwise}
  \end{cases}
\end{equation*}
In the example from Fig.~\ref{fig:scores}, setting $\tau=0.8$ results in \textit{Develop Hypothesis} and \textit{Develop Method} being flagged as normal, whereas all other activities are flagged as anomalous.

\subsection{Threshold Heuristic}
Most anomaly detection algorithms rely on the user setting a threshold manually or define the threshold as a constant.
To determine a threshold automatically, we propose a new heuristic that mimics how a human would set a threshold manually.

\begin{figure}[t]
  \centering
  \includegraphics[width=0.6\linewidth]{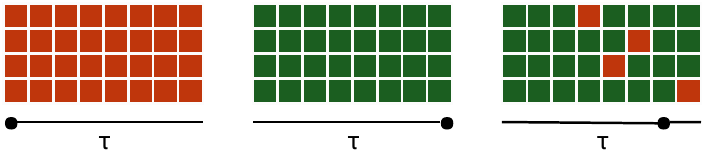}
  \caption{Example of how an anomaly detection visualization changes with different threshold settings; the rightmost setting corresponds to how a user would likely set the slider manually}
  \label{fig:threshold}
\end{figure}

Let us consider the following example.
A user is presented with a visualization of an anomaly detection result, say, a simple case overview showing all events and their attributes like depicted in Fig.~\ref{fig:threshold}.
Anomalous attributes are shown in red and normal attributes are shown in green.
The user is asked to set the threshold manually using a slider.
Most people start with the slider either set to the maximum (all attributes are normal, all green) or the minimum (all attributes are anomalous, all red) and then move the slider while observing the change of colors in the visualization.
Intuitively, most users fix the slider within a region where the number of shown anomalies is stable, that is, even when moving the slider to the left and right, the visualization stays the same.
Furthermore, users likely prefer a threshold setting that shows significantly less anomalous than normal attributes, which corresponds to a slider setting closer to the maximum (the right side).
In other words, a setting that produces less false positives.

This behavior of a human setting the threshold can be modeled based on the anomaly ratio $r$, which can be defined as follows, with $N = \sum_{c \in \mathcal{L}} |\hat{c}|$ denoting the number of non-padding events and $\vec{Y} = \theta(\vec{S}, \tau)$.
\begin{equation*}
  r(\theta, \vec{S}, \tau) = \frac{1}{NA} \sum_{i}^{C}\sum_{j}^{E}\sum_{k}^{A} \vec{Y}_{ijk}
\end{equation*}
By dividing by $N$ we calculate the average based only on non-padding events.

\begin{figure}[t]
  \centering
  \includegraphics[width=0.8\linewidth]{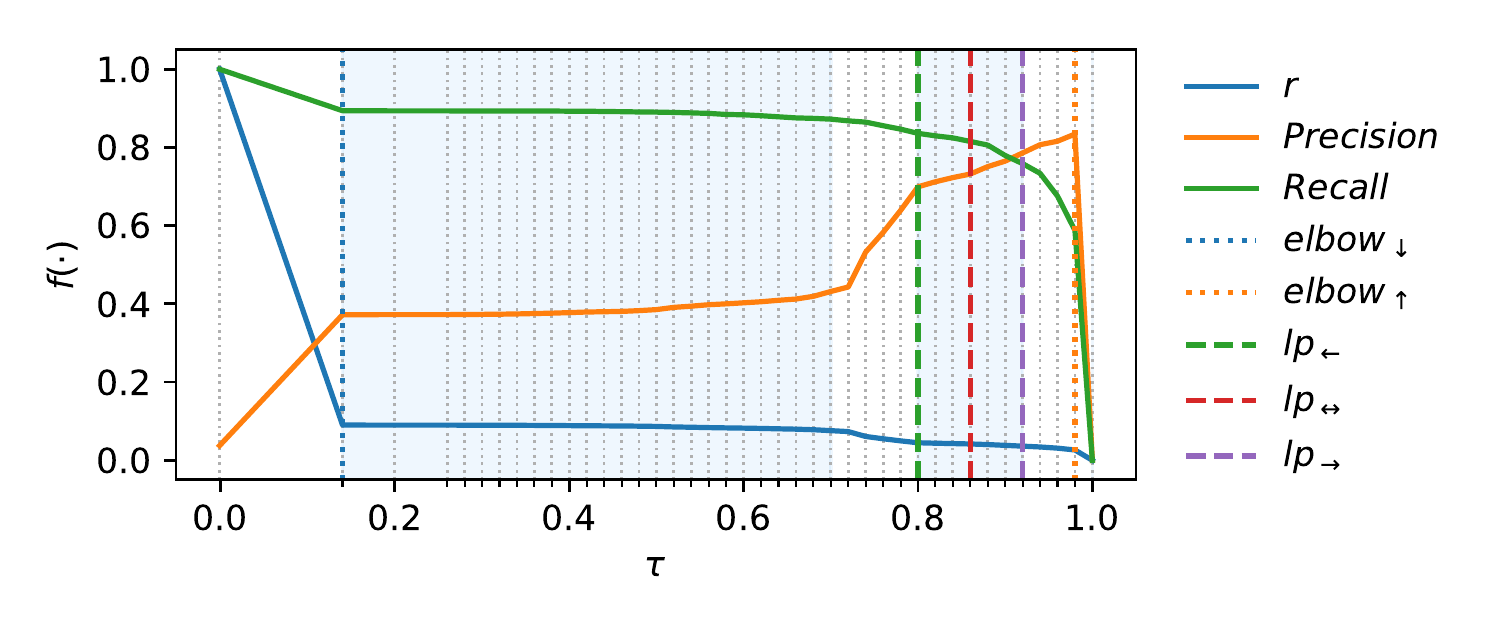}
  \caption{Thresholds as defined by the heuristics in relation to the anomaly ratio $r$ and its plateaus (blue intervals)}
  \label{fig:heuristics}
\end{figure}

Figure~\ref{fig:heuristics} shows $r$ for a run of BINetv1 on the Paper dataset.
In addition to $r$, the figure shows the values for $Precision$ and $Recall$ for the anomaly class.
Note that $r$ is a discrete function that we sample for each reasonable threshold.
Reasonable candidate thresholds are the distinct anomaly scores encountered in $\vec{S}$; other values can be disregarded.
The candidate thresholds $\mathcal{T}$ are indicated by the minor ticks and the respective dotted grid lines in Fig.~\ref{fig:heuristics}.

To define the heuristics in the following, we first have to define the first and second order derivatives of the discrete function $r$.
They can be retrieved using the central difference approximation. 
Let $\tau_i \in \mathcal{T}$, then the derivatives are approximated by
\begin{align*}
  r'(\theta, \vec{S}, \tau_i) &\approx \frac{r(\theta, \vec{S}, \tau_{i+1}) - r(\theta, \vec{S}, \tau_i)}{\tau_{i + 1} - \tau_{i}}, \\
  r''(\theta, \vec{S}, \tau_i) &\approx \frac{r(\theta, \vec{S}, \tau_{i-1}) - 2~r(\theta, \vec{S}, \tau_i) + r(\theta, \vec{S}, \tau_{i-1})}{(\tau_{i - 1} - \tau_{i})(\tau_{i} - \tau_{i+1})}.
\end{align*}

To mimic the human intuition, we have to consider regions of $r$ where the slope is close to zero.
These are regions where $|r'(\theta, \vec{S}, \tau)| < \varepsilon$ (we chose $\varepsilon$ to be two times the average slope of $r$), which we refer to as plateaus (blue regions in Fig.~\ref{fig:heuristics}).
Based on these plateaus we can now define the lowest plateau heuristic $lp$ as follows.
Let $LP = \langle\tau_0, \tau_1, \ldots, \tau_n\rangle$ be the sequence of candidate thresholds that lie within the lowest plateau, then
\begin{equation*}
  lp_\leftarrow = \tau_0, \quad
  lp_\leftrightarrow = \frac{1}{n} \sum_{i = 0}^{n} \tau_i, \quad
  lp_\rightarrow = \tau_n,
\end{equation*}
corresponding to the left-most ($lp_\leftarrow$), the right-most ($lp_\rightarrow$), and the mean-centered ($lp_\leftrightarrow$) threshold inside the lowest plateau.

In~\cite{nolle2018binet} we proposed the elbow heuristic to mimic the same behavior. 
The definition of the elbow heuristics are given by
\begin{equation*}
  elbow_\downarrow = \argmax_{\tau \in \mathcal{T}} r''(\theta, \vec{S}, \tau), \quad
  elbow_\uparrow = \argmin_{\tau \in \mathcal{T}} r''(\theta, \vec{S}, \tau).
\end{equation*}
So, $elbow_\downarrow$ is the threshold where the rate of change of $r$ is maximized, whereas $elbow_\uparrow$ is the minimum.
With respect to $r$ these thresholds are the points where either a steep drop ends in a plateau ($elbow_\downarrow$) or a plateau ends in a steep drop ($elbow_\uparrow$).
Although $elbow_\downarrow$ and $elbow_\uparrow$ can indicate the beginning and the end of a plateau, these are not necessarily the thresholds a human would naturally pick.

To compare our results to the best possible threshold, we define the $best$ heuristic by use of the $F_1$ score metric.
The $best$ heuristic is defined as follows, where $\vec{L}$ is the set of ground truth labels.
\begin{equation*}
  best = \argmax_{\tau \in \mathcal{T}} F_1(\vec{L}, \theta(\vec{S}, \tau)).
\end{equation*}
It is important to understand that $best$ can only be used if the labels are available at runtime.
However, in most cases, anomaly detection is an unsupervised problem, and hence no labels are available.

It might be beneficial to apply different thresholds to different dimensions of $\vec{S}$.
For example, it might be sensible to set a different threshold for the user attribute than the activity because the inherent probability distribution can be different.
This is possible by using ``:'' to apply heuristics on cross-sections of $\vec{S}$ by using index notation.
Let $h \in \{lp_\leftarrow, lp_\leftrightarrow, lp_\rightarrow, elbow_\downarrow, elbow_\uparrow, best\}$ then we can define the following threshold strategies
\begin{align*}
  h &= h(\vec{S}), \\
  h^{(a)} &= (\tau_i) = h(\vec{S}_{:,:,i}), \\
  h^{(e)} &= (\tau_i) = h(\vec{S}_{:,i,:}), \\
  h^{(ea)} &= (\tau_{ij}) = h(\vec{S}_{:,i,j}). \\
\end{align*}
We only explicitly show the parameter $\vec{S}$ for clarity, other parameters are set according to the definition of the chosen heuristic.

Thus, $h^{(a)} \in \mathbb{R}^A$ returns a tensor that holds one threshold for each attribute in an event, whereas $h^{(e)} \in \mathbb{R}^E$ holds a threshold for each event position in a case.
Lastly, $h^{(ea)} \in \mathbb{R}^{E \times A}$ combines the two ideas and gives a threshold for each combination of event position and attribute.
In other words, instead of applying the threshold heuristic $h$ once for all dimensions of $\vec{S}$, we apply it multiple times for different cross-sections of $\vec{S}$, obtaining multiple different thresholds.

\section{Evaluation}\label{sec:evaluation}
We evaluated BINet on all 44 event logs and compared it to eight state-of-the-art methods.
Two methods from~\cite{chandola2012survey}: a sliding window approach (t-STIDE+)~\cite{warrender1999detecting}; and the one-class SVM (OC-SVM).
Additionally, we compared BINet to two approaches from~\cite{bezerra2013algorithms}: the Naive algorithm and the Sampling algorithm.
Furthermore, we provide the results of the denoising autoencoder (DAE) approach from~\cite{nolle2018analyzing}.
Lastly, we compared BINet to the approach from~\cite{bohmer2016multi} (Likelihood).
Naive and Likelihood set the threshold automatically, so we extended the approaches to support the use of external threshold heuristics.
These extensions are referred to by Naive+ and Likelihood+.
For all non-deterministic methods (i.e., DAE, BINet, and Sampling), we executed five independent runs to counteract the randomness.

For the OC-SVM, we relied on the implementation of scikit-learn\footnote{http://scikit-learn.org} using an RBF kernel of degree 3 and $\nu = 0.5$.
The Naive, Sampling, Likelihood, and DAE methods were implemented as described in the original papers.
Sampling, Likelihood, Baseline, and the OC-SVM do not rely on a manual setting of the threshold and were unaltered.
t-STIDE+ is an implementation of the t-STIDE method from~\cite{warrender1999detecting}, which we adapted to support the data perspective (see~\cite{nolle2018analyzing}).
Naive+ is an implementation of Naive that removes the fixed threshold of $0.02$ and sets the threshold according to the heuristic.
Likelihood+ implements the first part of Likelihood (the generation of the extended likelihood graph from the log) and replaces the threshold algorithm with the aforementioned heuristics.

\begin{figure}[t]
  \centering
  \includegraphics[width=1.0\linewidth]{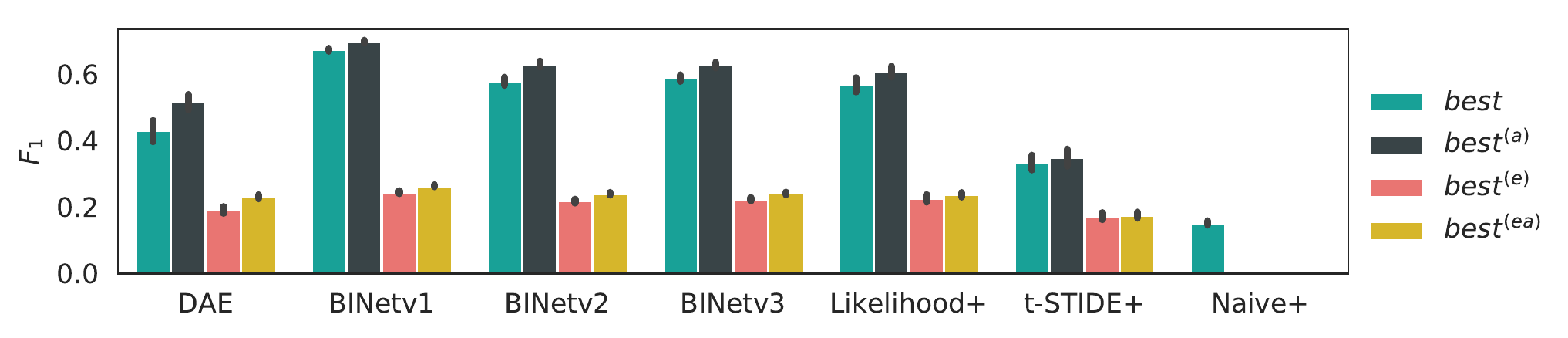}
  \caption{Average $F_1$ score by method and strategy over all synthetic datasets, using $best$ as the heuristic}
  \label{fig:eval_strategies}
\end{figure}

In the last section, we described the intuition of setting separate thresholds using different strategies (e.g., one threshold per attribute).
To decide on the best strategy, we evaluate the four strategies ($h$, $h^{(e)}$, $h^{(a)}$, and $h^{(ea)}$) for all synthetic datasets and all methods that support the heuristics, with $h = best$.
The results of the experiments in Fig.~\ref{fig:eval_strategies} indicate that, indeed, it is sensible to set separate thresholds for individual attributes.
Interestingly, we also find that setting a single threshold yields similar results.
Setting a threshold per event or per event and attribute does perform significantly worse.

\begin{figure}[t]
  \centering
  \includegraphics[width=1.0\linewidth]{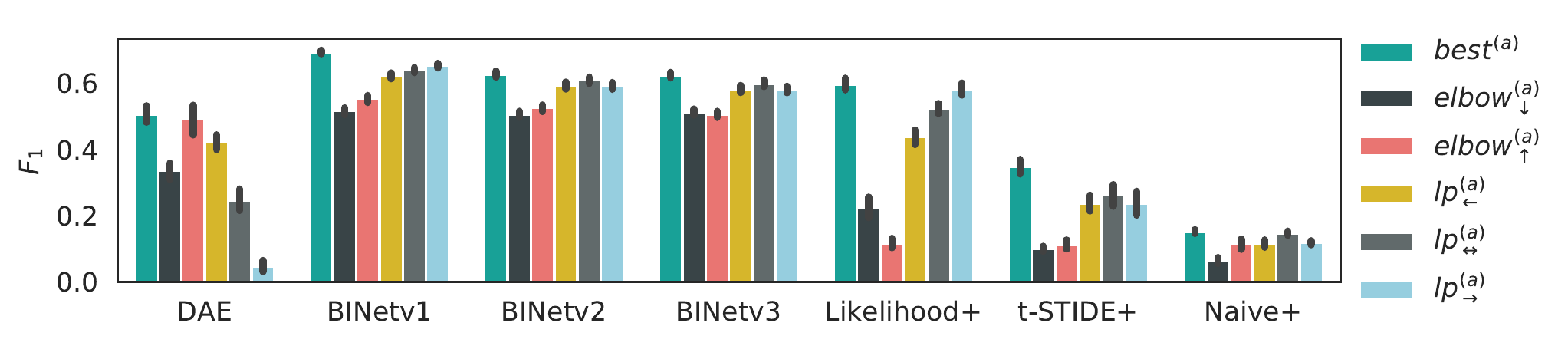}
  \caption{Average $F_1$ score by method and heuristic over all synthetic datasets, using $h^{(a)}$ as the strategy}
  \label{fig:eval_heuristics}
\end{figure}

Next, we repeated the same experiment for all of the aforementioned heuristics and using $h^{(a)}$ as the strategy.
The results can be seen in Fig.~\ref{fig:eval_heuristics}.
Intriguingly, the lowest plateau heuristics perform best for all methods except the DAE.
Furthermore, it seems to work best to choose the mean-centered threshold within the lowest plateau ($lp_\leftrightarrow$).

Based on the results of the preliminary experiments, we set $h = lp_\leftrightarrow^{(a)}$ as the heuristic for the following experiments for all methods apart from the DAE, for which we set $h = elbow_\uparrow^{(a)}$.
For Likelihood, Sampling, Naive, and OC-SVM we use the internal threshold heuristics.

\begin{figure}[t]
  \centering
  \includegraphics[width=1.0\linewidth]{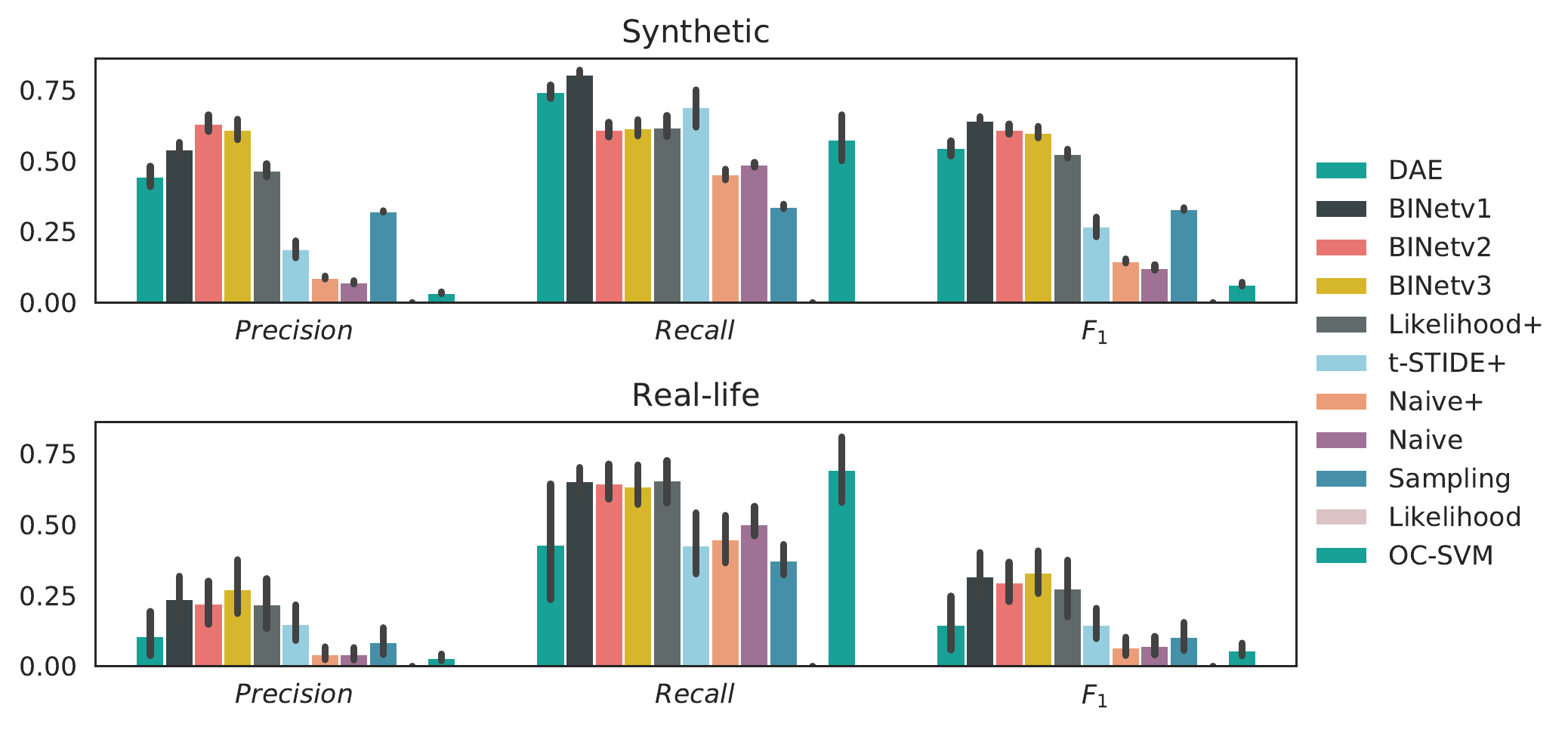}
  \caption{Average $Precision$, $Recall$, and $F_1$ by dataset type over all datasets; error bars indicate variance over datasets with different numbers of attributes and multiple runs}
  \label{fig:eval_overall}
\end{figure}

The overall results are shown in Fig.~\ref{fig:eval_overall}.
Note that for the real-life datasets we do not have complete information, and hence the $F_1$ score is not a good representation of the quality of the detection algorithms.
However, because we compare all methods on the same basis, the results are still meaningful.
Furthermore, we only know about the artificial anomalies inside the real-life datasets, and therefore we expect a high recall (of the artificial anomalies), whereas we expect a low precision because the dataset likely contains natural anomalies (which are not labeled).

This theory is confirmed by the results in Fig.~\ref{fig:eval_overall}.
Note also that the recall scores for both the synthetic and the real-life datasets are very similar, indicating comparable performance (for artificial anomalies) on both dataset types.

Finally, we find that BINetv1 works best on the synthetic datasets, whereas the field is mixed on the real-life datasets.
However, all three BINet versions perform better than the other methods.
DAE performs significantly worse on real-life data because it ran out of memory on some of the bigger datasets.
Therefore, DAE has been penalized defining precision and recall to be zero for these runs.

Detailed results can be found in Tab.~\ref{tab:results}, which also gives results for case level (i.e., only anomalous cases have to be detected, not the attributes).
An interesting observation is that t-STIDE+ performs best on BPIC12 when evaluating on case level.
This might be attributed to BPIC12 being a dataset without event attributes (the only one in the corpus).
On attribute level, Likelihood+ is marginally better than BINet on BPIC13.
For all other datasets, BINet shows the best performance.
\begin{table}[t]
  \centering
  \label{tab:results}
  \resizebox{\columnwidth}{!}{%
  \begin{tabular}{llccccccccccccc}
    \toprule
    Level     & Method & Paper & P2P & Small & Medium & Large & Huge & Gigantic & Wide & BPIC12 & BPIC13 & BPIC15 & BPIC17 & Anonymous \\
    \midrule
    Case      & OC-SVM \cite{warrender1999detecting}  & 0.49               & 0.27               & 0.25               & 0.29               & 0.24               & 0.23               & 0.29               & 0.31               & 0.55               & 0.24               & 0.26               & 0.35          & 0.10 \\
              & Naive \cite{Bezerra2009Anomaly}       & 0.50               & 0.48               & 0.49               & 0.39               & 0.41               & 0.40               & 0.34               & 0.44               & 0.55               & 0.21               & 0.17               & 0.31          & 0.16 \\
              & Sampling \cite{Bezerra2009Anomaly}    & 0.50               & 0.49               & 0.49               & 0.47               & 0.49               & 0.49               & 0.45               & 0.49               & 0.55               & 0.21               & 0.17               & 0.32          & 0.23 \\
              & Likelihood \cite{bohmer2016multi}.    & 0.00               & 0.00               & 0.00               & 0.00               & 0.00               & 0.00               & 0.00               & 0.00               & 0.00               & 0.00               & 0.00               & 0.00          & 0.00 \\
              & Naive+                                & 0.50               & 0.48               & 0.49               & 0.44               & 0.49               & 0.45               & 0.38               & 0.47               & 0.55               & 0.21               & 0.17               & 0.28          & 0.15 \\
              & t-STIDE+ \cite{nolle2016unsupervised} & 0.40               & 0.51               & 0.53               & 0.43               & 0.45               & 0.45               & 0.41               & 0.47               & \textbf{0.68}      & 0.32               & 0.29               & 0.32          & 0.22 \\
              & Likelihood+                           & \textbf{0.85}      & 0.74               & 0.76               & 0.72               & 0.73               & 0.73               & 0.73               & 0.73               & 0.62               & 0.44               & 0.33               & 0.45          & 0.51 \\
              & DAE \cite{nolle2016unsupervised}      & 0.46               & 0.71               & 0.72               & 0.71               & 0.71               & 0.70               & 0.63               & 0.70               & 0.60               & 0.21               & 0.00               & 0.30          & 0.35 \\
              & BINetv1                               & 0.74               & \textbf{0.77}      & \textbf{0.78}      & \textbf{0.75}      & \textbf{0.75}      & \textbf{0.75}      & \textbf{0.74}      & \textbf{0.76}      & 0.62               & 0.41               & 0.37               & \textbf{0.51} & \textbf{0.51} \\
              & BINetv2 \cite{nolle2018binet}         & 0.76               & 0.77               & 0.77               & 0.72               & 0.71               & 0.70               & 0.68               & 0.73               & 0.61               & 0.40               & \textbf{0.38}      & 0.43          & 0.45 \\
              & BINetv3                               & 0.79               & 0.77               & 0.76               & 0.71               & 0.69               & 0.69               & 0.66               & 0.74               & 0.66               & \textbf{0.45}      & 0.36               & 0.49          & 0.50 \\
    \midrule
    Attribute & OC-SVM \cite{warrender1999detecting}  & 0.09               & 0.06               & 0.05               & 0.08               & 0.04               & 0.05               & 0.07               & 0.09               & 0.05               & 0.06               & 0.01               & 0.09          & 0.30 \\
              & Naive \cite{Bezerra2009Anomaly}       & 0.13               & 0.15               & 0.14               & 0.12               & 0.09               & 0.11               & 0.09               & 0.16               & 0.05               & 0.05               & 0.01               & 0.10          & 0.39 \\
              & Sampling \cite{Bezerra2009Anomaly}    & 0.33               & 0.33               & 0.34               & 0.32               & 0.34               & 0.34               & 0.31               & 0.32               & 0.08               & 0.07               & 0.01               & 0.14          & 0.39 \\
              & Likelihood \cite{bohmer2016multi}     & 0.00               & 0.00               & 0.00               & 0.00               & 0.00               & 0.00               & 0.00               & 0.00               & 0.00               & 0.00               & 0.00               & 0.00          & 0.00 \\
              & Naive+                                & 0.13               & 0.16               & 0.15               & 0.16               & 0.13               & 0.13               & 0.13               & 0.18               & 0.05               & 0.05               & 0.01               & 0.09          & 0.33 \\
              & t-STIDE+ \cite{nolle2016unsupervised} & 0.28               & 0.33               & 0.32               & 0.25               & 0.25               & 0.26               & 0.19               & 0.28               & 0.40               & 0.12               & 0.05               & 0.17          & 0.39 \\
              & Likelihood+                           & \textbf{0.74}      & 0.61               & 0.63               & 0.58               & 0.57               & 0.58               & 0.55               & 0.57               & 0.35               & 0.29               & 0.07               & 0.28          & 0.63 \\
              & DAE \cite{nolle2016unsupervised}      & 0.25               & 0.61               & 0.61               & 0.56               & 0.56               & 0.55               & 0.46               & 0.56               & 0.06               & 0.09               & 0.00               & 0.24          & 0.52 \\
              & BINetv1                               & 0.64               & \textbf{0.68}      & \textbf{0.69}      & \textbf{0.65}      & \textbf{0.65}      & \textbf{0.65}      & \textbf{0.64}      & \textbf{0.65}      & 0.42               & 0.28               & \textbf{0.21}      & \textbf{0.38} & 0.60 \\
              & BINetv2 \cite{nolle2018binet}         & 0.67               & 0.65               & 0.67               & 0.60               & 0.59               & 0.60               & 0.56               & 0.60               & 0.34               & 0.25               & 0.19               & 0.29          & 0.55 \\
              & BINetv3                               & 0.67               & 0.65               & 0.66               & 0.59               & 0.57               & 0.59               & 0.54               & 0.61               & \textbf{0.48}      & \textbf{0.29}      & 0.19               & 0.35          & \textbf{0.63} \\
    \bottomrule
  \end{tabular}
  }
  \caption{$F_1$ score over all datasets by detection level and method; best results (before rounding) are shown in bold typeface}
\end{table}

All results are given using the heuristics described above.
Labels were not used in the process.
Additional material (e.g., evaluation per perspective, per dataset, runtime) can be found in the respective code repository.

\begin{figure}[t]
  \centering
  \includegraphics[width=0.6\linewidth]{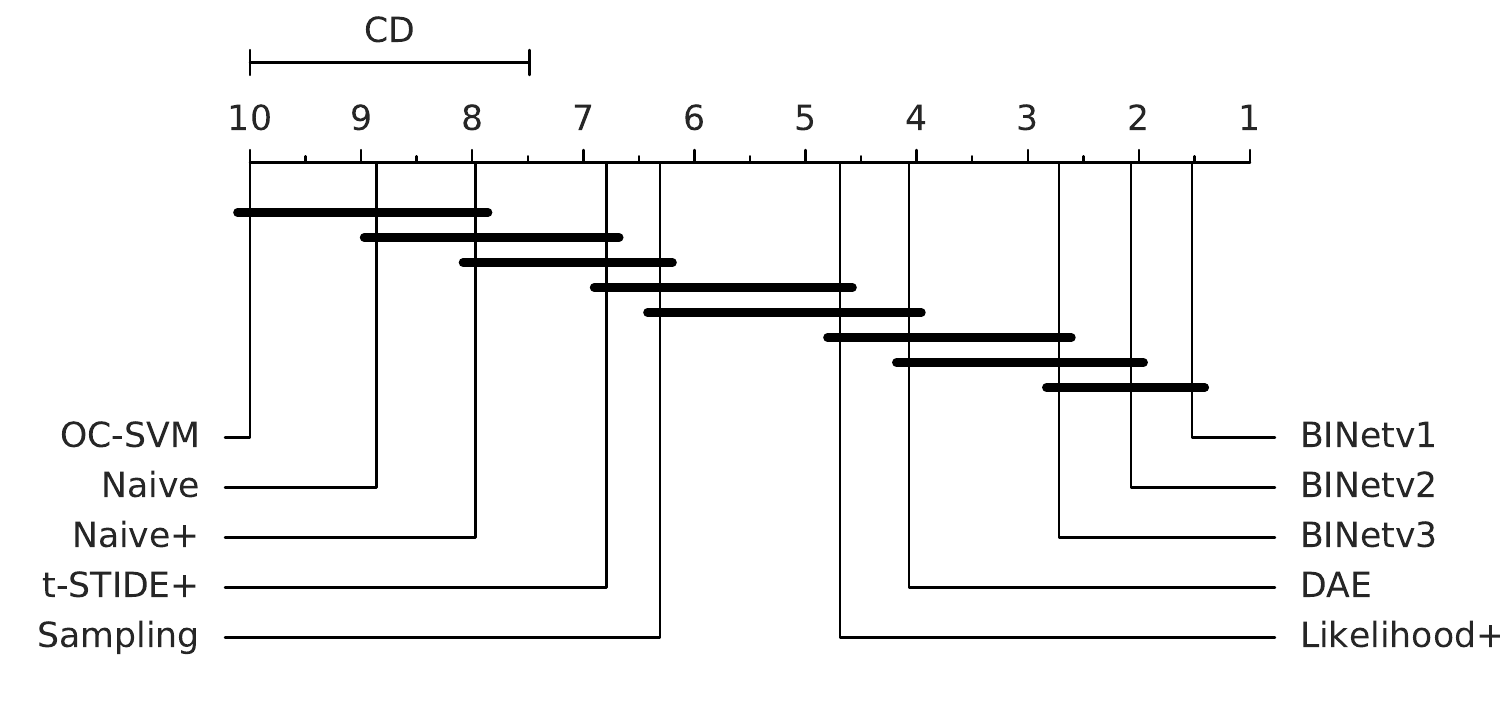}
  \caption{Critical difference diagram for all methods on all synthetic datasets; groups of methods that are not significantly different (at $p = 0.05$) are connected (cf.~\cite{demvsar2006statistical})}
  \label{fig:significance}
\end{figure}

To validate the significance of the results, we apply the non-parametric Friedman test~\cite{friedman1937use} on average ranks of all methods based on $F_1$ score for all synthetic datasets.
Then, we apply the Nemenyi post-hoc test~\cite{nemenyi1963dist} as demonstrated in~\cite{demvsar2006statistical}, to calculate pairwise significance.
Figure~\ref{fig:significance} shows a critical difference (CD) diagram as proposed in~\cite{demvsar2006statistical} to visualize the results with a confidence interval of 95 percent.
Based on the critical difference, we recognize that BINetv1 performs significantly better than all other methods, except BINetv2 and BINetv3.
That is, all three BINet versions lie in the same significance group with respect to the critical difference.
DAE lies in the same group as BINetv2 and BINetv3, and Likelihood+ in the same as DAE and BINetv3.
All other methods lie more than the critical difference away from the three BINets.

\section{Classifying Anomalies}
\label{sec:classification}
Until now, we have not utilized the predictive capabilities that BINet possesses.
Using the probability distribution output of the softmax layers in conjunction with the binarized anomaly scores, we can define simple rules to infer the type (or class) of an anomaly.

We use the term predictions to denote all possible attribute values that lie within the confidence interval determined by the threshold.
Note that now the \textit{Shift} class becomes relevant because it indicates the place where an early or late execution would belong.
For each anomalous attribute (according to BINet), we apply the following rules in order.
\begin{enumerate}
  \item \textit{Skip}: If all predictions do not appear somewhere in the case
  \item \textit{Insert}: If one of the predictions appears somewhere in the case and that occurrence has not been flagged as anomalous
  \item \textit{Rework}: If the same activity is present somewhere earlier in the case and is not flagged as anomalous
  \item \textit{Shift}: If one of the prediction appears either somewhere earlier or later in the case and is flagged as anomalous
  \item \textit{Late}: If the activity appears somewhere earlier in the predictions and is flagged as anomalous
  \item \textit{Early}: If the activity appears somewhere later in the predictions and is flagged as anomalous
  \item \textit{Attribute}: Trivially, all anomalous attributes in the data perspective are of type \textit{Attribute}
\end{enumerate}

\begin{figure}[t]
  \centering
  \includegraphics[width=1.0\linewidth]{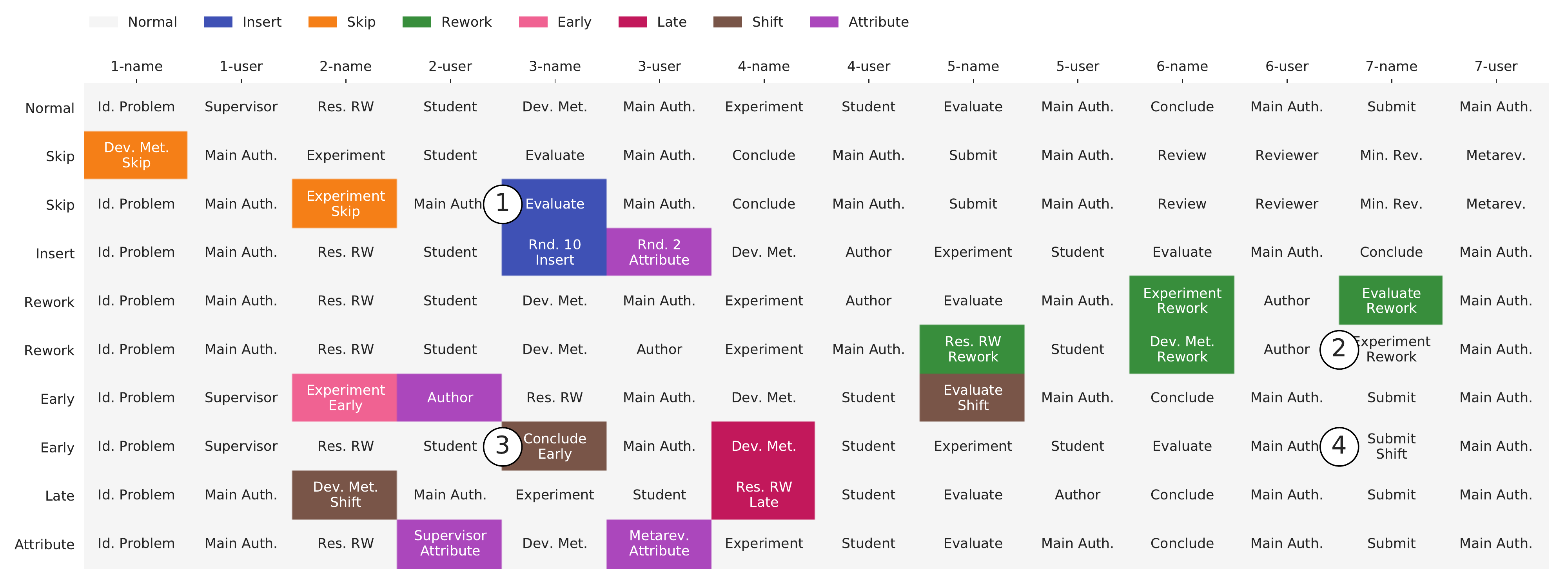}
  \caption{Classification of anomalies on the Paper dataset based on anomaly scores from BINetv1 using $h = lp_\rightarrow^{(a)}$; colors indicate the prediction of the classifier (see legend) and actual classes are shown as text within the cells}
  \label{fig:classification}
\end{figure}

The result of the classification is visualized in Fig.~\ref{fig:classification}.
Interestingly, this set of simple rules performs remarkably well.
Anomaly classes inferred by the rules are indicated by the color of the cells, whereas ground truth labels are shown as text in the cells (we omit \textit{Normal} for clarity).
Incidentally, this visualization also depicts the binarized anomaly scores according to the threshold since each classified attribute is also an anomalous attribute.

We also included some examples where the classification is incorrect.
An interesting case is the second \textit{Skip} example because \textit{Evaluate} has also been marked as anomalous \ding{172}.
As we have defined in Fig.~\ref{fig:likelihood}, \textit{Evaluate} is an activity that always eventually follows \textit{Develop Method}.
However, \textit{Develop Method} was skipped.
Therefore, BINet is never presented with the causing activity, and hence regards \textit{Evaluate} as anomalous.

A different example is that BINet misses the third \textit{Rework} activity in the second example \ding{173}.
We observed many of these errors, and they are related to the fact that BINet is conditioned on the last input activity and forgets the history of the case (forgetting problem).
Under these conditions, \textit{Develop Method} indeed directly follows \textit{Research Related Work}, and hence BINet misses it.
This forgetting problem is something we want to address in the future.

The most interesting case is the second \textit{Early} example \ding{174}.
Here, BINet misclassifies the \textit{Early} activity as \textit{Shift}.
Upon closer inspection, we realize that this is indeed a way of explaining the anomaly, albeit not the one the labels indicate.
With respect to \textit{Develop Method}, \textit{Conclude} indeed occurs too early in the case.
Nevertheless, BINet fails to detect the actual Shift point \ding{175}, and hence the rules do not match the pattern correctly.

\begin{figure}[t]
  \centering
  \includegraphics[width=0.8\linewidth]{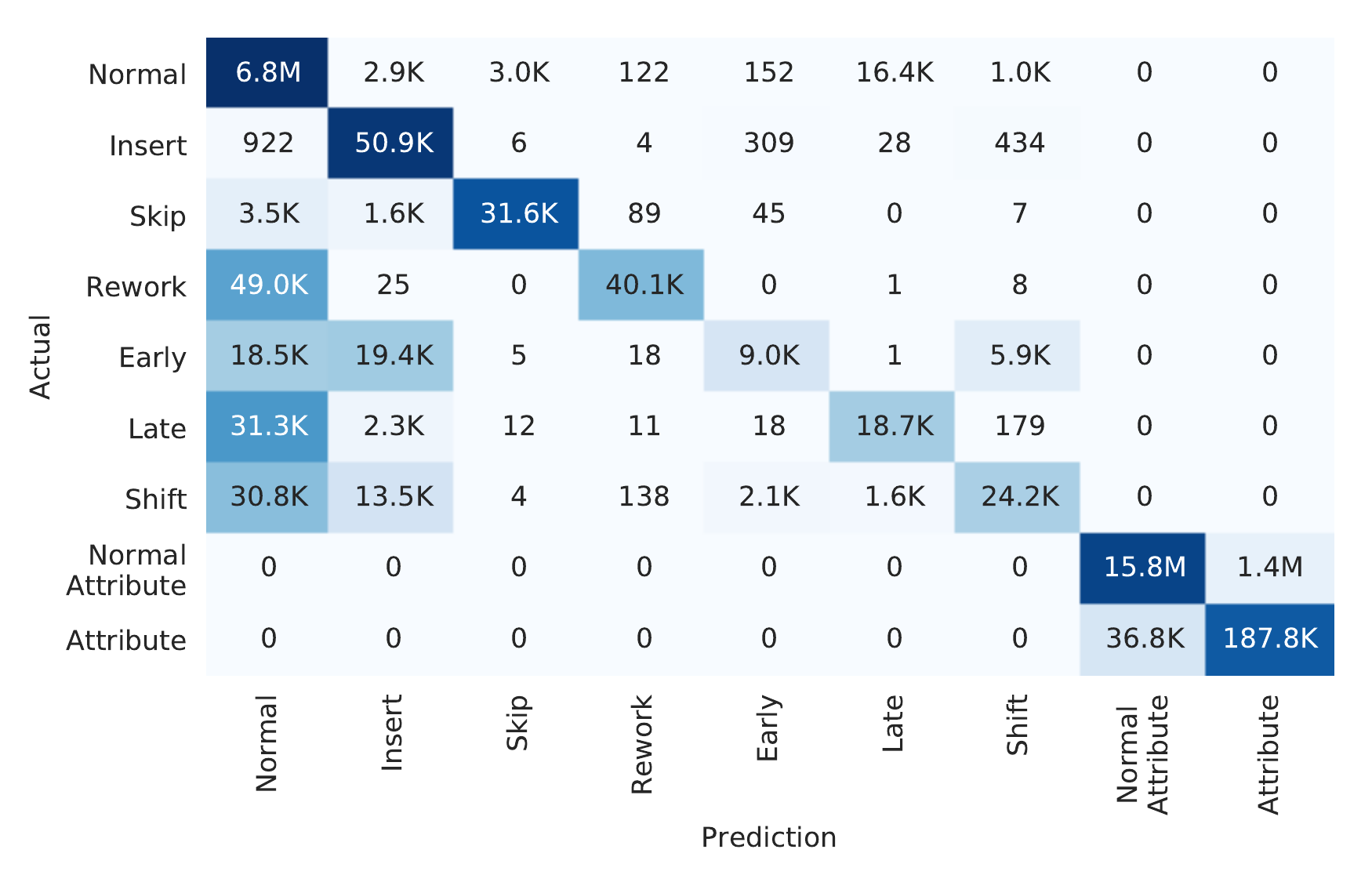}
  \caption{Confusion matrix for all runs of BINetv1 on synthetic datasets with  $h = lp_\rightarrow^{(a)}$; color indicates distribution of actual class}
  \label{fig:cm}
\end{figure}

Using this simple set of rules we ran the classifier on all synthetic datasets, using BINetv1 as the anomaly detection method and $h = lp_\rightarrow^{(a)}$ (the best heuristic for BINetv1).
Figure~\ref{fig:cm} shows the results in a confusion matrix.
Note that the classifier uses as input the anomaly detection result of BINet, and hence can never distinguish normal from anomalous examples.
Thus, the errors for the normal class are based on the errors BINet commits in the anomaly detection task.
Disregarding these errors, this results in a macro average $F_1$ score of $0.83$ over all datasets for the classification task.
Since BINetv1 reaches an average $F_1$ score of $0.64$ on the detection task, this result is truly impressive, considering the simplicity of the rules.
For the joint task (detection and classification), BINetv1 reaches an average $F_1$ score of $0.70$.

In Fig.~\ref{fig:cm}, we notice that BINet errs especially often for \textit{Rework}, \textit{Early}, \textit{Late}, and \textit{Shift}.
This is connected to the forgetting problem mentioned earlier.
Remember that \textit{Rework}, \textit{Early}, and \textit{Late} anomalies are affecting sequences of events, that is, up to 3 events can be part of a rework anomaly, and up to 2 events can be executed early or late.
In the case of Rework, we have already seen an example in Fig.~\ref{fig:classification}, where BINet misclassifies because of forgetting.
Figure~\ref{fig:cm} confirms that this error occurs quite often since more than 50 percent of all  \textit{Rework} anomalies are misclassified as \textit{Normal}.
All of these misclassifications happen in cases where a sequence of more than one event has been executed again.

Note that not every repetition of an event is classified as a \textit{Rework}, only the events identified to be anomalous are.
Hence, a repeated event (a loop in the process) is classified as \textit{Normal}, if BINet has learned that it can occur multiple times in a case.
In the Paper process, this is demonstrated by the second \textit{Submit} event, which can naturally occur multiple times in a case.
In Fig.~\ref{fig:cm} we can see that BINet very rarely classifies a \textit{Normal} activity as \textit{Rework} (never in the Paper datasets); thus, we can conclude that BINet has learned to model the loop in the Paper process correctly.

As with \textit{Rework}, we can explain the errors for \textit{Early} and \textit{Late} by the same argument.
However, these two classes are also often misclassified as \textit{Insert} or \textit{Shift}.
The latter goes back to the second \textit{Early} example of Fig.~\ref{fig:classification} and the ambiguity of labels.
The \textit{Insert} errors are of a different kind.
They occur because the rule set is not taking into account that multiple events can be executed early or late.
We expect to find an early execution somewhere later in the case as the prediction; however, this can only be true for the first event of an early sequence.
The same argument can be made for late executions.

The \textit{Shift} errors are related to the fact that the random process models often allow skipping of events.
When a \textit{Shift} anomaly is applied to an optional event, BINet, or any other method, has no means of finding the anomaly.
This could be accounted for by altering the generation algorithm.

Nevertheless, the results indicate that a simple set of rules can be used to classify the anomalies types we have introduced before.
Note that this is a white-box approach and a human user can easily interpret the resulting classification.
Even though the different classes are only a subset of all anomaly types, they do cover many of the anomalies encountered in real-life business processes.
Importantly, it is quite easy to define new rules for new types of anomalies based on the predictive capabilities of BINet.

\section{Conclusion}\label{sec:conclusion}
In this paper, we presented three versions of BINet, a neural network architecture for multi-perspective anomaly classification in business process event logs.
Additionally, we proposed a set of heuristics for setting the threshold of an anomaly detection algorithm automatically, based on the anomaly ratio function.
Finally, we demonstrated that a simple set of rules could be used for classification of anomaly types, solely based on the output of BINet.

BINet is a recurrent neural network, and can, therefore, be used for real-time anomaly detection, since it does not require a completed case for detection.
BINet does not rely on any information about the process modeled by an event log, nor does it depend on a clean dataset.
Utilizing the lowest plateau heuristic, BINet's internal threshold can be set automatically, reducing manual workload and allowing fully autonomous operation.

It can be used to find point anomalies as well as contextual anomalies because it models the sequential nature of the cases utilizing both the control flow and the data perspective.
Furthermore, BINet can cope with concept drift, for it can be set up to train on new cases in real-time continuously.

Based on the empirical evidence obtained in the evaluation, BINet is a promising method for anomaly detection, especially in business process event logs.
BINet outperformed the opposition on all detection levels.
Specifically, on the synthetic datasets, BINet's performance surpasses those of other methods by an order of magnitude.
We demonstrated that BINet also performs well on the real-life datasets because BINet shows high recall of the artificial anomalies introduced to the original real-life logs.

Even though the results look very promising, there is still room for improvement.
For example, BINet suffers from forgetting when sequences of events are repeated in a case.
This issue can be addressed in future work, for example, by using a special attention layer.
An interesting option is the use of a bidirectional encoder-decoder structure to read in cases both from left to right and from right to left.
Hereby, sequences of repeated events can be identified from two sides, as opposed to just one.

Overall, the results presented in this paper suggest that BINet is a reliable and versatile method for detecting---and classifying---anomalies in business process logs.

\section*{Acknowledgments}\label{sec:Acknowledgments}
This project [522/17-04] is funded in the framework of Hessen ModellProjekte, financed with funds of LOEWE, F\"orderlinie 3: KMU-Verbundvorhaben (State Offensive for the Development of Scientific and Economic Excellence), and by the German Federal Ministry of Education and Research (BMBF) Software Campus project ``AI.RPM'' [01IS17050].

\section*{References}
\bibliography{references}

\end{document}